\documentclass{article}

\usepackage[nonatbib]{arxiv}

\usepackage[utf8]{inputenc} 
\usepackage[T1]{fontenc}    
\usepackage{hyperref}       
\usepackage{endnotes}
\usepackage{url}            
\usepackage{booktabs}       
\usepackage{amsfonts}       
\usepackage{nicefrac}       
\usepackage{microtype}      
\usepackage{graphicx}
\usepackage{float}
\usepackage{amsmath}
\usepackage{amssymb}
\usepackage[square,sort,comma,numbers]{natbib}

\DeclareMathOperator{\median}{median}

\title{From free text to clusters of content in health records: An unsupervised graph partitioning approach}

\author{
  M. Tarik ~Altuncu \\
  Department of Mathematics\\
  Imperial College London\\
  London, SW7 2AZ \\
  \texttt{m.altuncu16@imperial.ac.uk} \\
\And
  Erik ~Mayer \\
  Department of Surgery \& Cancer\\
  Imperial College London\\
  London, SW7 2AZ \\
  \texttt{e.mayer@imperial.ac.uk} \\
\And
  Sophia N. ~Yaliraki \\
  Department of Chemistry\\
  Imperial College London\\
  London, SW7 2AZ \\
  \texttt{s.yaliraki@imperial.ac.uk} \\
\And
  Mauricio ~Barahona \\
  Department of Mathematics\\
  Imperial College London\\
  London, SW7 2AZ \\
  \texttt{m.barahona@imperial.ac.uk} \\
}

\begin{document}
\maketitle

\begin{abstract}
Electronic Healthcare records contain large volumes of unstructured data in different forms. Free text constitutes a large portion of such data, yet this source of richly detailed information often remains under-used in practice because of a lack of suitable methodologies to extract interpretable content in a timely manner. Here we apply network-theoretical tools to the analysis of free text in Hospital Patient Incident reports in the English National Health Service, to find clusters of reports in an unsupervised manner and at different levels of resolution based directly on the free text descriptions contained within them. To do so, we combine recently developed deep neural network text-embedding methodologies based on paragraph vectors with multi-scale Markov Stability community detection applied to a similarity graph of documents obtained from sparsified text vector similarities. We showcase the approach with the analysis of incident reports submitted in Imperial College Healthcare NHS Trust, London. The multiscale community structure reveals levels of meaning with different resolution in the topics of the dataset, as shown by relevant descriptive terms extracted from the groups of records, as well as by comparing \textit{a posteriori} against hand-coded categories assigned by healthcare personnel. Our content communities exhibit good correspondence with well-defined hand-coded categories, yet our results also provide further medical detail in certain areas as well as revealing complementary descriptors of incidents beyond the external classification. 
We also discuss how the method can be used to monitor reports over time and across different healthcare providers, and to detect emerging trends that fall outside of pre-existing categories.
\end{abstract}



\section*{Introduction}

The vast amounts of data collected by healthcare providers in conjunction with modern data analytics techniques present a unique opportunity to improve health service provision and the quality and safety of medical care for patient benefit~\citep{colijn2017toward}. Much of the recent research in this area has been on personalised medicine and its aim to deliver better diagnostics aided by the integration of diverse datasets providing complementary information. Another large source of healthcare data is organisational. In the United Kingdom, the National Health Service (NHS) has a long history of documenting extensively the different aspects of healthcare provision. The NHS is currently in the process of increasing the availability of several databases, properly anonymised, with the aim of leveraging advanced analytics to identify areas of improvement in NHS services.

One such database is the National Reporting and Learning System (NRLS), a central repository of patient safety incident reports from the NHS in England and Wales. Set up in 2003, the NRLS now contains more than 13 million detailed records. The incidents are reported using a set of standardised categories and contain a wealth of organisational and spatio-temporal information (structured data), as well as, crucially, a substantial component of free text (unstructured data) where incidents are described in the `voice' of the person reporting. The incidents are wide ranging: from patient accidents to lost forms or referrals; from delays in admission and discharge to serious untoward incidents, such as retained foreign objects after operations. The review and analysis of such data provides critical insight into the complex functioning of different processes and procedures in healthcare towards service improvement for safer carer.

Although statistical analyses are routinely performed on the structured component of the data (dates, locations, assigned categories, etc), the free text remains largely unused in systematic processes. 
Free text is usually read manually but this is time-consuming, meaning that it is often ignored in practice, unless a detailed review of a case is undertaken because of the severity of harm that resulted. There is a lack of methodologies that can summarise content and provide content-based groupings across the large volume of reports submitted nationally for organisational learning.  
Methods that could provide automatic categorisation of incidents from the free text would sidestep problems such as difficulties in assigning an incident category by virtue of \textit{a priori} pre-defined lists in the reporting system or human error, as well as offering a unique insight into the root cause analysis of incidents that could improve the safety and quality of care and efficiency of healthcare services.

Our goal in this work is to showcase an algorithmic methodology that detects content-based groups of records in a given dataset in an unsupervised manner, based only on the free and unstructured textual description of the incidents.  
To do so, we combine recently developed deep neural-network high-dimensional text-embedding algorithms with network-theoretical methods.  In particular, we apply multiscale Markov Stability (MS) community detection to a sparsified geometric similarity graph of documents obtained from text vector similarities. 
Our method departs from traditional natural language processing tools, which have generally used bag-of-words (BoW) representation of documents and statistical methods based on Latent Dirichlet Allocation (LDA) to cluster documents~\citep{lda}.
More recent approaches have used deep neural network based language models clustered with k-means, without a full multiscale graph analysis~\citep{Hashimoto2016TopicReviews}. 
There have been some previous applications of network theory to text analysis. For example, Lanchichinetti and co-workers 
~\citep{PhysRevX.5.011007} used a probabilistic graph construction analysed with the InfoMap algorithm~\citep{infomap_EPJS}; however, their community detection was carried out at a single-scale and the representation of text as BoW arrays lacks the power of neural network text embeddings. 
The application of multiscale community detection allows us to find groups of records with consistent content at different levels of resolution; hence the content categories emerge from the textual data, rather than fitting with pre-designed classifications. 
The obtained results could thus help mitigate possible human error or effort in finding the right category in complex category classification trees.

We showcase the methodology through the analysis of a dataset of patient incidents reported to the NRLS. First, we use the 13 million records collected by the NRLS since 2004 to train our text embedding (although a much smaller corpus can be used). We then analyse a subset of 3229 records reported from St Mary's Hospital, London (Imperial College Healthcare NHS Trust) over three months in 2014 to extract clusters of incidents at different levels of resolution in terms of content. 
Our method reveals multiple levels of intrinsic structure in the topics of the dataset, 
as shown by the extraction of relevant word descriptors from the grouped records and a high level of topic coherence.
Originally, the records had been manually coded by the operator upon reporting with up to 170 features per case, including a two-level manual classification of the incidents. 
Therefore, we also carried out an \textit{a posteriori} comparison against the hand-coded categories assigned by the reporter (healthcare personnel) at the time of the report submission. Our results show good overall correspondence with the hand-coded categories across resolutions and, specifically, at the medium level of granularity. Several of our clusters of content correspond strongly to well-defined categories, yet our results also reveal complementary categories of incidents not defined in the external classification. In addition, the tuning of the granularity afforded by the method can be used to provide a distinct level of resolution in certain areas corresponding to specialise or particular sub-themes.

\section*{Multiscale graph partitioning for text analysis: description of the framework}

Our framework combines text-embedding, geometric graph construction and multi-resolution community detection to identify, rather than impose, content-based clusters from free, unstructured text in an unsupervised manner.

\begin{figure}[htb]
\includegraphics[width=.9\linewidth]{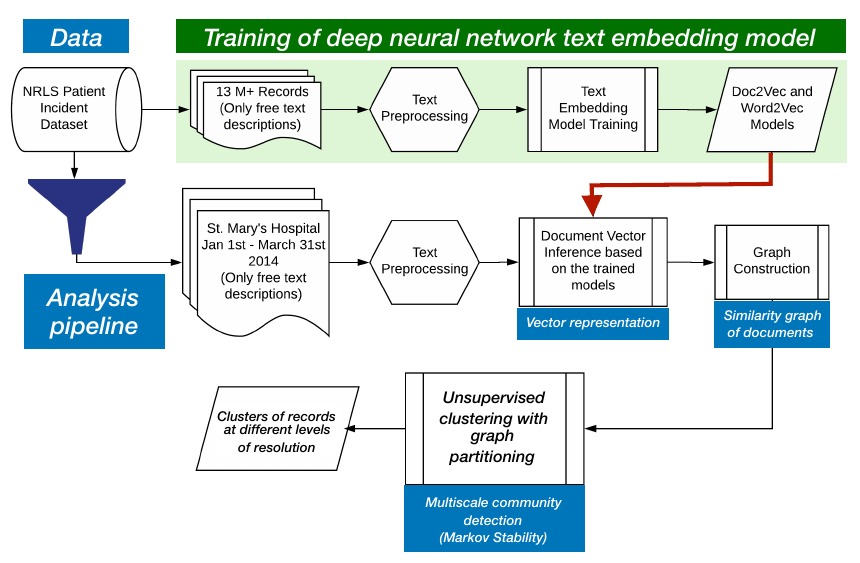}
\caption{
Pipeline for data analysis including the training of the text embedding model and the graph-based unsupervised clustering of documents at different levels of resolution to find topic clusters only from the free text descriptions of hospital incident reports from the NRLS database.}
\label{fig:pipeline}
\end{figure}

Figure \ref{fig:pipeline} shows a summary of our pipeline. First, we pre-process each document to transform text into consecutive word tokens, where words are in their most normalised forms, and some words are removed if they have no distinctive meaning when used out of context~\citep{nltk,porter_old}. 
We then train a paragraph vector model using the Document to Vector (Doc2Vec) framework~\citep{d2v_mikolov} on the whole set (13 million) of preprocessed text records, although training on smaller sets (1 million) also produces good results. This training step is only done once.
This Doc2Vec model is subsequently used to infer high-dimensional vector descriptions for the text of each of the 3229 documents in our target analysis set. 
We then compute a matrix containing pairwise similarities between any pair of document vectors, as inferred with Doc2Vec.
This matrix can be thought of as a full, weighted graph with documents as nodes and edges weighted by their similarity.  We sparsify this graph to the union of a minimum spanning tree and a k-Nearest Neighbors (MST-kNN) graph~\citep{mstknn}, a geometric construction that removes less important similarities but preserves global connectivity for the graph and, hence, for the dataset. 
The derived MST-kNN graph is analysed with Markov Stability~\citep{pnasStability,LambiotteMarkovProcess,Delvenne2013,lambiotte_arxiv}, a multi-resolution dynamics-based graph partitioning method that identifies relevant subgraphs (i.e., clusters of documents) at different levels of granularity. MS uses a diffusive process on the graph to reveal the multiscale organisation at different resolutions without the need for choosing {\it a priori} the number of clusters, scale or organisation. 
To analyse {\it a posteriori} the different partitions across levels of resolution, we use both visualisations and quantitative scores. The visualisations include word clouds to summarise the main content, graph layouts, as well as Sankey diagrams and contingency tables that capture the correspondences across levels of resolution and relationships to the hand-coded classifications. The partitions are also evaluated quantitatively to score: (i) their intrinsic topic coherence (using pairwise mutual information~\citep{pmi_coherence, pmi_coherence2}), and (ii) their similarity to the operator hand-coded categories (using normalised mutual information~~\citep{nmi}). We now expand on the steps of the computational framework.

\subsection*{Data description} 

The full dataset includes more than 13 million confidential reports of patient safety incidents reported to the \textit{National Reporting and Learning System} (NRLS) between 2004 and 2016 from NHS trusts and hospitals in England and Wales. Each record has more than 170 features, including organisational details (e.g., time, trust code and location), anonymised patient information, medication and medical devices, among other details.
The records are manually classified by operators to a two-level system of categories of incident type. In particular, the top level contains 15 categories including general groups such as `Patient accident', `Medication', `Clinical assessment', `Documentation', `Admissions/Transfer' or `Infrastructure' alongside more specific groups such as `Aggressive behaviour', `Patient abuse', `Self-harm' or `Infection control'.
In most records, there is also a detailed description of the incident in free text, although the quality of the text is highly variable. 
Our analysis set for clustering is the group of 3229 records reported during the first quarter of 2014 at St. Mary's Hospital in London (Imperial College Healthcare NHS Trust). 

\subsection*{Text Preprocessing} 

Text preprocessing is important to enhance the performance of text embedding. We applied standard preprocessing techniques in natural language processing to the raw text of all 13 million records in our corpus. We normalise words into a single form and remove words that do not carry significant meaning. Specifically, we divide our documents into iterative word tokens using the NLTK library~\citep{nltk} and remove punctuation and digit-only tokens. We then apply word stemming using the Porter algorithm~\citep{porter_old,porter}. If the Porter method cannot find a stemmed version for a token, we apply the Snowball algorithm~\citep{snowball}. Finally, we remove any stop-words (repeat words with low content) using NLTK's stop-word list. Although some of the syntactic information is reduced due to text preprocessing, this process preserves and consolidates the semantic information of the vocabulary, which is of relevance to our study. 
\label{textPreprocessing}

\subsection*{Text Embedding} 

Computational methods for text analysis rely on a choice of a mathematical representation of the base units, such as character $n$-grams, words or documents of any length. 
An important consideration for our methodology is an attempt to avoid the use of labelled data at the core of many supervised or semi-supervised classification methods~\citep{semevalSTS2016, semevalSTS2017}.
In this work, we use a representation of text documents in vector form following recent developments in the field. 

Classically, bag-of-words (Bow) methods were used to obtain representations of the documents in a corpus in terms of vectors of term frequencies weighted by inverse document frequency (TF-iDF).
While such methods provide a statistical description of documents, they do not carry information about the order or proximity of words to each other since they regard word tokens in an independent manner with no semantic or syntactic relationships considered.
Furthermore, BoW representations tend to be high-dimensional and sparse, due to large sizes of word dictionaries and low frequencies of many terms.

Recently, deep neural network language models have successfully overcome certain limitations of BoW methods 
by incorporating word neighbourhoods in the mathematical description of each term.  
Distributed Bag of Words (DBOW) is a form of Paragraph Vectors (PV), also known as Doc2Vec~\citep{d2v_mikolov}. This method creates a model which represents any length of word sequences (i.e. sentences, paragraphs, documents) as $d$-dimensional vectors, where $d$ is a user-defined parameter (typically $d=500$).
Training a Doc2Vec model starts with a random $d$-dimensional vector assignment for each document in the corpus. A stochastic gradient descent algorithm iterates over the corpus with the objective of predicting a randomly sampled set of words from each document by using only the document's $d$-dimensional vector~\citep{d2v_mikolov}.
The objective function being optimised by PV-DBOW is similar to the skip-gram model in Refs.~\citep{mikolov2013efficient,w2v2}.
Doc2Vec has been shown~\citep{dai2015document} to capture both semantic and syntactic characterisations of the input text outperforming BoW models, such as LDA~\citep{lda}.

Here, we use the Gensim Python library~\citep{gensim} to train the PV-DBOW model.
The Doc2Vec training was repeated several times with a variety of training hyper-parameters to optimise the output based on our own numerical experiments and the general guidelines provided by~\citep{jhlau}.
We trained Doc2Vec models using text corpora of different sizes and content with different sets of hyper-parameters, in order to characterise the usability and quality of models. 
Specifically, we checked the effect of corpus size on model quality by training Doc2Vec models on the full 13 million NRLS records and on subsets of 1 million and 2 million randomly sampled records. (We note that our target subset of 3229 records has been excluded from these samples.) Furthermore, we checked the importance of the specificity of the text corpus by obtaining a Doc2Vec model from a generic, non-specific set of 5 million articles from Wikipedia representing standard English usage across a variety of topics.

\paragraph*{Benchmarking of the Doc2Vec training.}
We benchmarked the Doc2Vec models by scoring how well the document vectors represent the semantic topic structure:
(i) calculating centroids for the 15 externally hand-coded categories; 
(ii) selecting the 100 nearest reports for each centroid;
(iii) counting the number of incident reports (out of 1500) correctly assigned to their centroid. 
The results in Table~\ref{table:d2v} show that training on the highly specific text in the NRLS records is an important ingredient in the successful vectorisation of the documents, as shown by the degraded performance for the Wikipedia model across a variety of training hyper-parameters. 
Our results also show that reducing the size of the corpus from 13 million to 1 million records did not affect the benchmarking dramatically. This robustness of the results to the size of the training corpus was confirmed further with the use of more detailed metrics, as discussed below in Section~\nameref{sec:comparisons}.

\begin{table}[htb]
\centering
\begin{tabular}{|c|c|c||c|c|c|c|}
\hline
\multicolumn{3}{|c||}{\textbf{Hyper-parameters}} & \multicolumn{3}{c|}{\textbf{NRLS}} & \textbf{Wikipedia} \\ \hline
\textbf{\begin{tabular}[c]{@{}c@{}}Window \\ Size\end{tabular}} & \textbf{\begin{tabular}[c]{@{}c@{}}Minimum \\ Count\end{tabular}} & \textbf{Subsampling} & \textbf{1M} & \textbf{2M} & \textbf{13M+} & \textbf{5M+} \\ \hline
15 & 5 & 0.001 & 765 & 755 & \textbf{836} & 531 \\ \hline
5 & 5 & 0.001 & 807 & 775 & 798 & 580 \\ \hline
5 & 20 & 0.001 & 801 & 785 & 809 & 587 \\ \hline
5 & 20 & 0.00001 & - & - & 379 & 465 \\ \hline
15 & 20 & 0.00001 & - & - & 387 & 424 \\ \hline 
\end{tabular}
\bigskip
\caption{Benchmarking of text corpora used for Doc2Vec training. A Doc2Vec model was trained on three corpora of NRLS records of different sizes and a corpus of Wikipedia articles using a variety of hyper-parameters. The scores represent the quality of the vectors inferred using the corresponding model: the number of correct assignments out of 1500.}
\label{table:d2v}
\end{table}

Based on our benchmarking, we use henceforth (unless otherwise noted) the optimised Doc2Vec model obtained from the 13+ million NRLS records with the following hyper-parameters: \{training method = dbow, number of dimensions for feature vectors size = 300, number of epochs = 10, window size = 15, minimum count = 5, number of negative samples = 5, random down-sampling threshold for frequent words = 0.001 \}. As an indication of computational cost, the training of the model on the 13 million records takes approximately 11 hours (run in parallel with 7 threads) on shared servers.

\subsection*{Graph Construction}

Once the Doc2Vec model is trained, we use it to infer a vector for each of the $N=3229$ records in our analysis set. We then construct a normalised cosine similarity matrix between the vectors by: computing the matrix of cosine similarities between all pairs of records, $S_\text{cos}$; transforming it into a distance matrix $D_{cos} = 1-S_{cos}$; applying element-wise max norm to obtain $\hat{D}=\|D_{cos}\|_{max}$; and normalising the similarity matrix $\hat{S} = 1-\hat{D}$ which has elements in the interval $[0,1]$.

\begin{figure}[htb]
\includegraphics[width=.97\linewidth]{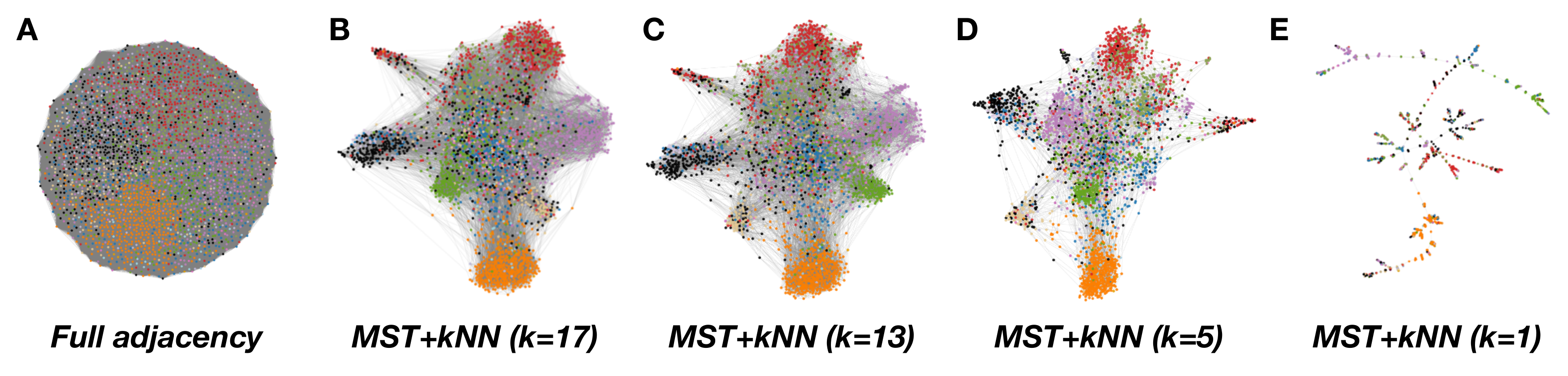}
\caption{
Planar layouts using the ForceAtlas2 algorithm~\citep{forceAtlas2} of some of the similarity graphs generated from the dataset of 3229 records. Each node represents a record and is coloured according to its hand-coded, external category to aid visualisation of the structure. Note that the external categories are not used to produce our content-driven multi-resolution clustering in Figure~\ref{fig:MS}. 
\textbf{(a)} Layout for the full, weighted normalised similarity matrix $\hat{S}$ without MST-kNN applied. 
\textbf{(b)}--\textbf{(e)} Layouts of the graphs generated from the data with the MST-kNN algorithm with an increasing level of sparsity: $k=17, 13, 5, 1$ respectively. The structure of the graph is sharpened for intermediate values of $k$, and we choose $k=13$ for our analysis here. }
\label{fig:fa2_graphConstruction}
\end{figure}

The similarity matrix can be thought of as the adjacency matrix of a fully connected weighted graph. However, such a graph contains many edges with small weights reflecting weak similarities in 
high-dimensional noisy datasets even the least similar nodes present a substantial degree of similarity. Such weak similarities are in most cases redundant, as they can be explained through stronger pairwise similarities present in the graph.  These weak, redundant edges obscure the graph structure, as shown by the diffuse, spherical visualisation of the full graph layout in Figure~\ref{fig:fa2_graphConstruction}A.

To reveal the graph structure, we obtain a MST-kNN graph from the normalised similarity matrix~\citep{mstknn}. This is a simple sparsification based on a geometric heuristic that preserves the global connectivity of the graph while retaining details about the local geometry of the dataset. 
The MST-kNN algorithm starts by computing the minimum spanning tree (MST) of the full matrix $\hat{D}$, i.e., the tree with $(N-1)$ edges connecting all nodes in the graph with minimal sum of edge weights (distances). The MST is computed using the Kruskal algorithm implemented in SciPy~\citep{scipy}. 
To this MST, we add edges connecting each node to its $k$ nearest nodes (kNN) if they are not already in the MST. Here $k$ is an user-defined parameter. 
The binary adjacency matrix of the MST-kNN graphs, $E_\text{MST-kNN}$, is Hadamard-multiplied with $\hat{S}$ to give the adjacency matrix $A$ of the weighted, undirected sparsified graph.
The MST-kNN method avoids a direct thresholding of the weights in $\hat{S}$, and obtains a graph description that preserves local geometric information together with a global subgraph (the MST) that captures properties of the full dataset.

The network layout visualisations in Figure~\ref{fig:fa2_graphConstruction}B--E give an intuitive picture of the effect of the sparsification.
The highly sparse graphs obtained when the number of neighbours $k$ is very small are not robust. As $k$ is increased, the local similarities between documents induce the formation of dense subgraphs (which appear closer in the graph visualisation layout). When the number of neighbours becomes too large, the local structure becomes diffuse and the subgraphs lose coherence, signalling the degradation of the local graph structure. 
Figure~\ref{fig:fa2_graphConstruction} shows that the MST-kNN graph with $k=13$ presents a reasonable balance between local and global structure. 
Relatively sparse graphs that preserve important edges and global connectivity of the dataset (guaranteed here by the MST) have computational advantages when using community detection algorithms.

The MST-kNN construction has been reported to be robust to the selection of the parameter $k$ due to the guaranteed connectivity provided by the MST~\citep{mstknn}. 
In the following, we fix $k=13$ for our analysis with the multi-scale graph partitioning framework, but we have scanned values of $k \in [1,50]$ in the graph construction from our data and have found that the construction is robust as long as $k$ is nor too small (i.e., $k > 13$). The detailed comparisons are shown in Section~\nameref{sec:comparisons}.

The MST-kNN construction has the advantage of its simplicity and robustness, and the fact that it balances the local and global structure of the data. 
However, the area of network inference and graph construction from data, and graph sparsification is very active, and several alternative approaches exist based on different heuristics, e.g., Graphical Lasso~\citep{g_lasso}, Planar Maximally Filtered Graph~\citep{pmfg}, spectral sparsification~\citep{spielman2011graph}, or the Relaxed Minimum Spanning Tree (RMST)~\citep{rbs_rmst}. 
We have experimented with some of those methods and obtained comparable results. A detailed comparison of sparsification methods as well as the choice of distance in defining the similarity matrix $\hat S$ is left for future work.

\subsection*{Multiscale Graph Partitioning}

The area of community detection encompasses a variety of graph partitioning approaches which aim to find `good' partitions into subgraphs (or communities) according to different cost functions, without imposing the number of communities \textit{a priori}~\citep{Schaub2017}. The notion of community thus depends on the choice of cost function. Commonly, communities are taken to be subgraphs whose nodes are connected strongly within the community with relatively weak inter-community edges. Such structural notion is related to balanced cuts. Other cost functions are posed in terms of transitions inside and outside of the communities, usually as one-step processes~\citep{infomap_EPJS}. When transition paths of random walks of all lengths are considered, the concept of community becomes intrinsically multi-scale, i.e., different partitions can be found to be relevant at different time scales leading to a multi-level description dictated by the transition dynamics~\citep{pnasStability,Schaub2012ZoomingLens,LambiotteMarkovProcess}. This leads to the framework of Markov Stability, a dynamics-based, multi-scale community detection methodology, which can be shown to recover seamlessly several well-known heuristics as particular cases~\citep{pnasStability,Delvenne2013,lambiotte_arxiv}.

Here, we apply MS to find partitions of the similarity graph $A$ at different levels of resolution. The subgraphs detected correspond to clusters of documents with similar content.
MS is an unsupervised community detection method that finds robust and stable partitions under the evolution of a continuous-time diffusion process without {\it a priori} choice of the number or type of communities or their organisation~\citep{pnasStability,Schaub2012ZoomingLens,LambiotteMarkovProcess,ukRiotsTw}~\endnote{The code for Markov Stability is open and accessible at \url{https://github.com/michaelschaub/PartitionStability} and \url{http://wwwf.imperial.ac.uk/~mpbara/Partition_Stability/}, last accessed on March 24, 2018}. In simple terms, MS can be understood by analogy to a drop of ink diffusing on the graph under a diffusive Markov process. The ink diffuses homogeneously unless the graph has some intrinsic structural organisation, in which case the ink gets transiently contained, over particular time scales, within groups of nodes (i.e., subgraphs or communities). The existence of this transient containment signals the presence of a natural partition of the graph. As the process evolves, the ink diffuses out of those initial communities but might get transiently contained in other, larger subgraphs.
By analysing this Markov dynamics over time, MS detects the structure of the graph across scales. The Markov time $t$ thus acts as a resolution parameter that allows us to extract robust partitions that persist over particular time scales, in an unsupervised manner.

Given the adjacency matrix $A_{N \times N}$ of the graph obtained as described previously, let us define the diagonal matrix $D=\text{diag}(\mathbf{d})$, 
where $\mathbf{d}=A \mathbf{1}$ is the degree vector.
The random walk Laplacian matrix is defined as $L_\text{RW}=I_N-D^{-1}A$ where $I_N$ is the identity matrix of size $N$, and the transition matrix (or kernel) of the associated continuous-time Markov process is $P(t)=e^{-t L_\text{RW}}, \, t>0$~\citep{LambiotteMarkovProcess}.
For each partition, a binary membership matrix $H_{N \times C}$  maps the $N$ nodes into $C$ clusters.
We can then define the $C\times C$ clustered autocovariance matrix:
\begin{align}
R(t,H) = H^T[\Pi P(t)-\pi\pi^T]H \label{eq:diffusion} 
\end{align}
where $\pi$ is the steady-state distribution of the process and $\Pi=\text{diag}(\pi)$. The element $[R(t,H)]_{\alpha \beta}$ quantifies the probability that a random walker starting from community $\alpha$ will end in community $\beta$ at time $t$, subtracting the probability that the same event occurs by chance at stationarity.

We then define our cost function measuring the goodness of a partition over time $t$, termed the Markov Stability of partition $H$:
\begin{equation}
r(t,H) = \text{trace} \left[R(t,H)\right]. \label{eq:MS}
\end{equation}
A partition $H$ that maximises $r(t,H)$ is comprised of communities that preserve the flow within themselves over time $t$, since in that case the diagonal elements of $R(t,H)$ will be large and the off-diagonal elements will be small. 
For details, see ~\citep{pnasStability,Schaub2012ZoomingLens,LambiotteMarkovProcess,bacik_celegans}.

MS searches for partitions at each Markov time that maximise $r(t,H)$.
Although the maximisation of~\eqref{eq:MS} is an NP-hard problem (hence with no guarantees for global optimality), there are efficient optimisation methods that work well in practice. Our implementation here uses the Louvain Algorithm~\citep{louvain,lambiotte_arxiv} which is efficient and known to give good results when applied to benchmarks~\citep{Lancichinetti2009CommDetectCompare}. To obtain robust partitions, we run the Louvain algorithm 500 times with different initialisations at each Markov time and pick the best 50 with the highest Markov Stability value $r(t,H)$.  We then compute the variation of information~\citep{Meila2007} of this ensemble of solutions $VI(t)$. as a measure of the reproducibility of the result under the optimisation. In addition, the relevant partitions are required to be persistent across time, as given by low values of the variation of information between optimised partitions across time $VI(t,t')$. 
Robust partitions are thus indicated by Markov times where $VI(t)$ shows a dip and $VI(t,t')$ has an extended plateau, indicating consistent results from different Louvain runs and validity over extended scales~\citep{bacik_celegans,LambiotteMarkovProcess}. 

\subsection*{Visualisation and interpretation of the results} 

\paragraph*{Graph layouts.}
We use the ForceAtlas2~\citep{forceAtlas2} layout to represent the graph of 3229 NRLS Patient Incident reports. 
This layout follows a force-directed iterative method to find node positions that balance attractive and repulsive forces. Hence similar nodes tend to be grouped together on the planar layout. We colour the nodes by either hand-coded categories (Figure~\ref{fig:fa2_graphConstruction}) or multiscale MS communities (Figure~\ref{fig:MS}). Spatially consistent colourings on this layout imply good clusters of documents in terms of the similarity graph.

\paragraph*{Tracking membership through Sankey diagrams.}
Sankey diagrams allow us to visualise the relationship of node memberships across different partitions and with respect to the hand-coded categories.
In particular, two-layer Sankey diagrams (e.g., Fig.~\ref{fig:44comms}) reflect the correspondence between MS clusters and the hand-coded external categories, whereas the multilayer Sankey diagram in Fig.~\ref{fig:MS} represents the results of the multi-resolution MS community detection across scales.

\paragraph*{Normalised contingency tables.}
In addition to Sankey diagrams between our MS clusters and the hand-coded categories, we also provide a complementary visualisation as heatmaps of normalised contingency (z-score) tables, e.g., Fig.~\ref{fig:44comms}. This allows us to compare the relative association of content clusters to the external categories at different resolution levels. A quantification of this correspondence is also provided by the $NMI$ score introduced in Eq.~\eqref{eq:nmi}.

\paragraph*{Word clouds of increased intelligibility through lemmatisation.} 
Our method clusters text documents according to their intrinsic content. This can be understood as a type of topic detection. To understand the content of the clusters, we use Word Clouds as basic, yet intuitive, tools that summarise information from a group of documents. Word clouds allow us to evaluate the results and extract insights when comparing \textit{a posteriori} with hand-coded categories. They can also provide an aid for monitoring results when used by practitioners.

The stemming methods described in the \textit{Text Preprocessing} subsection truncate words severely. Such truncation enhances the power of the language processing computational methods, as it reduces the redundancy in the word corpus. Yet when presenting the results back to a human observer, it is desirable to report the content of the clusters with words that are readily comprehensible.  To generate comprehensible word clouds in our \textit{a posteriori} analyses, we use a text processing method similar to the one described in~\citep{wordClouds}.
Specifically, we use the part of speech (POS) tagging module from NLTK to leave out sentence parts except the adjectives, nouns, and verbs. 
We also remove less meaningful common verbs such as `be', `have', and `do' and their variations.
The residual words are then lemmatised and represented with their lemmas in order to normalise variations of the same word.
Once the text is processed in this manner, we use the Python library wordcloud\endnote{The word cloud generator library for Python is open and accessible at \url{https://github.com/amueller/word\_cloud}, last accessed on March 25, 2018} to create word clouds with 2 or 3-gram frequency list of common word groups. The results present distinct, understandable word topics.

\subsection*{Quantitative benchmarking of topic clusters}

Although our dataset has attached a hand-coded classification by a human operator, we do not use it in our analysis and we do not consider it as a `ground truth'. Indeed, one of our aims is to explore the relevance of the fixed external classes as compared to the content-driven groupings obtained in an unsupervised manner. Hence we provide a double route to quantify the quality of the clusters by computing two complementary measures:  an intrinsic measure of topic coherence and a measure of similarity to the external hand-coded categories, defined as follows.

\paragraph*{Topic coherence of text:} 
As an \textit{intrinsic} measure of consistency of word association without any reference to an external `ground truth', we use the \textit{pointwise mutual information} ($PMI$)~\citep{pmi_coherence, pmi_coherence2}. The $PMI$ is an information-theoretical score that captures the probability of being used together in the same group of documents. The $PMI$ score for a pair of words $(w_1,w_2)$ is:
\begin{equation}
PMI(w_1,w_2)=\log{\frac{P(w_1 w_2)}{P(w_1)P(w_2)}}
\label{eq:pmi}
\end{equation}
where the probabilities of the words $P(w_1)$, $P(w_2)$, and of their co-occurrence $P(w_1 w_2)$ are obtained from the corpus.
To obtain the aggregate $\widehat{PMI}$ for the graph partition $C=\{c_i\}$ we compute the $PMI$ for each cluster, as the median $PMI$ between its 10 most common words (changing the number of words gives similar results), and we obtain the weighted average of the $PMI$ cluster scores:
\begin{align}
\widehat{PMI} (C) = \sum_{c_i \in C} \frac{n_i}{N} \, \mathop{\median}_{\substack{w_k, w_\ell \in S_i \\ k<\ell}}  PMI(w_k,w_\ell),
\label{eq:pmi_partition}
\end{align}
where $c_i$ denotes the clusters in partition $C$, each with size $n_i$; $N=\sum_{c_i \in C} n_i$ is the total number of nodes; and $S_i$ denotes the set of top 10 words for cluster $c_i$.

We use this $\widehat{PMI}$ score to evaluate partitions without requiring a labelled ground truth.The $PMI$ score has been shown to perform well~\citep{pmi_coherence, pmi_coherence2} when compared to human interpretation of topics on different corpora~\citep{pmi_coherence_lda, twitter_pmi_coherence}, and is designed to evaluate topical coherence for groups of documents, in contrast to other tools aimed at short forms of text. See~\citep{semevalSTS2016, semevalSTS2017, semeval2016Samsung, semaval2017ECNU} for other examples.

\paragraph*{Similarity between the obtained partitions and the hand-coded categories:} 
To compare against the external classification \emph{a posteriori}, we use the normalised mutual information ($NMI$), a well-used information-theoretical score that quantifies the similarity between clusterings considering both the correct and incorrect assignments in terms of the information (or predictability) between the clusterings.  The NMI between two partitions $C$ and $D$ of the same graph is:
\begin{equation}
\label{eq:nmi}
NMI(C,D)=\frac{I(C,D)}{\sqrt{H(C)H(D)}}=\frac{\sum\limits_{c \in C} \sum\limits_{d \in D} p(c,d) \, \log\dfrac{p(c,d)}{p(c)p(d)}}{\sqrt{H(C)H(D)}}
\end{equation}
where $I(C,D)$ is the Mutual Information and $H(C)$ and $H(D)$ are the entropies of the two partitions.

The $NMI$ is bounded ($0 \leq NMI \leq 1$) with a higher value corresponding to higher similarity of the partitions (i.e., $NMI=1$ when there is perfect agreement between partitions $C$ and $D$). The $NMI$ score is directly related\endnote{\url{http://scikit-learn.org/stable/modules/generated/sklearn.metrics.v_measure_score.html}} to the \textit{V-measure} used in the computer science literature~\citep{vmeasure}.  
We use the $NMI$ to compare the partitions obtained by MS (and other methods) against the hand-coded classification assigned by the operator.

\section*{Application to the analysis of hospital incident reports} 

\subsection*{Multi-resolution community detection extracts content clusters at different levels of granularity}

\begin{figure}[htb]
\includegraphics[width=0.97\linewidth]{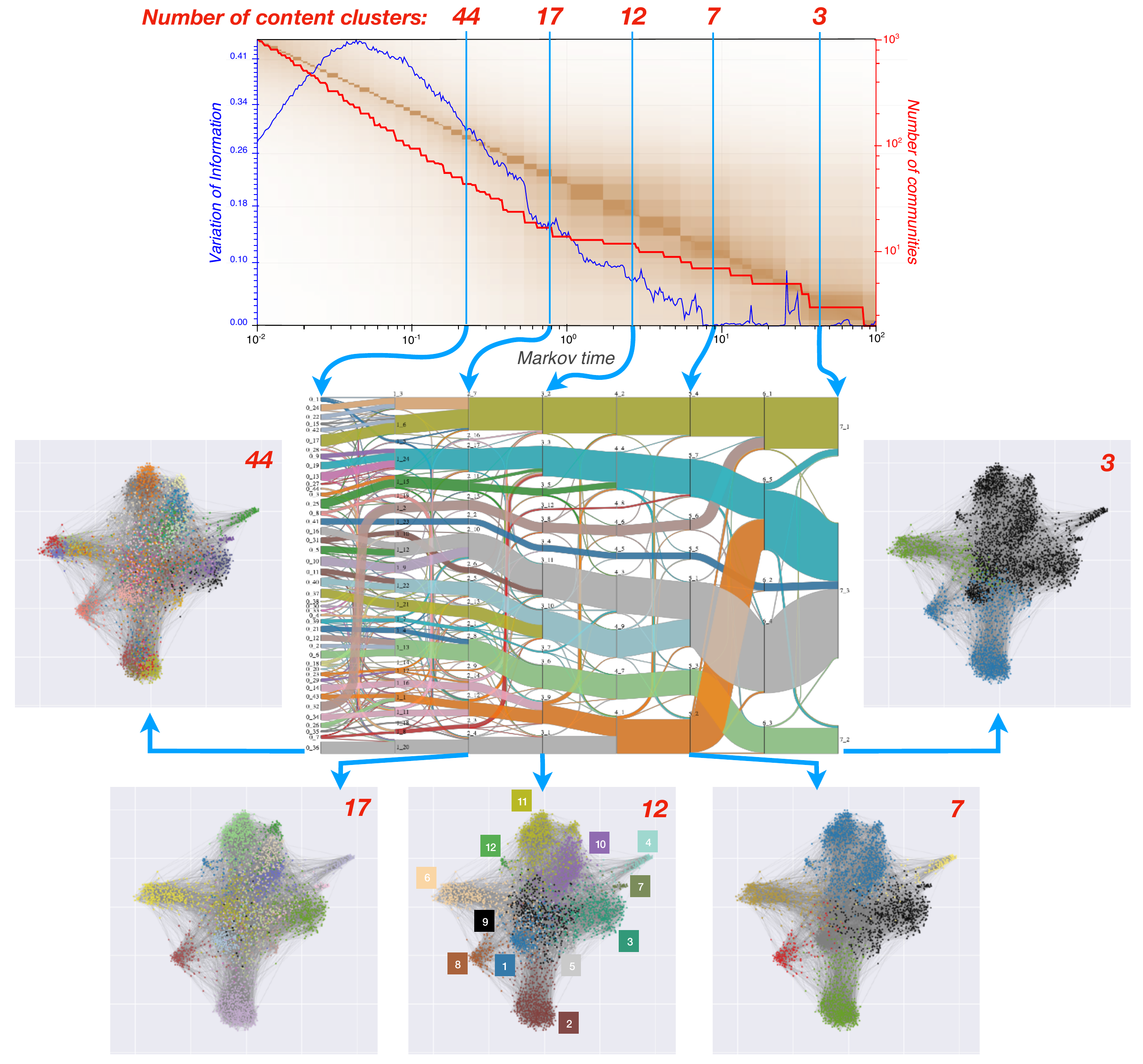}
\caption{
The top plot presents the results of the Markov Stability algorithm across Markov times, showing the number of clusters of the optimised partition (red), the variation of information $VI(t)$ for the ensemble of optimised solutions at each time (blue) and the variation of Information $VI(t,t')$ between the optimised partitions across Markov time (background colourmap). Relevant partitions are indicated by dips of $VI(t)$ and extended plateaux of $VI(t,t')$. We choose five levels with different resolutions (from 44 communities to 3) in our analysis. The Sankey diagram below illustrates how the communities of documents (indicated by numbers and colours) map across Markov time scales. The community structure across scales present a strong quasi-hierarchical character---a result of the analysis and the properties of the data, since it is not imposed \textit{a priori}. The different partitions for the five chosen levels are shown on a graph layout for the document similarity graph created with the MST-kNN algorithm with $k=13$. The colours correspond to the communities found by MS indicating content clusters.}
\label{fig:MS}
\end{figure}

We applied MS across a broad span of Markov times ($t \in [0.01, 100]$ in steps of 0.01) to the MST-kNN similarity graph of $N=3229$ incident records. At each Markov time, we ran 500 independent optimisations of the Louvain algorithm and selected the optimal partition at each time. Repeating the optimisation from 500 different initial starting points enhances the robustness of the outcome and allows us to quantify the robustness of the partition to the optimisation procedure. To quantify this robustness, we computed the average variation of information $VI(t)$ (a measure of dissimilarity) between the top 50 partitions for each $t$. Once the full scan across Markov time was finalised, a final comparison of all the optimal partitions obtained was carried out, so as to assess if any of the optimised partitions was optimal at any other Markov time, in which case it was selected. We then obtained the $VI(t,t')$ across all optimal partitions found across Markov times to ascertain when partitions are robust across levels of resolution. This layered process of optimisation enhances the robustness of the outcome given the NP-hard nature of MS optimisation, which prevents guaranteed global optimality.

Figure~\ref{fig:MS} presents a summary of our analysis. We plot the number of clusters of the optimal partition and the two metrics of variation of information across all Markov times. The existence of a long plateau in $VI(t,t')$ coupled to a dip in $VI(t)$ implies the presence of a partition that is robust both to the optimisation and across Markov time. To illustrate the multi-scale features of the method, we choose several of these robust partitions, from finer (44 communities) to coarser (3 communities), obtained at five Markov times and examine their structure and content. We also present a multi-level Sankey diagram to summarise the relationships and relative node membership across the levels.

The MS analysis of the graph of incident reports reveals a rich multi-level structure of partitions, with a strong quasi-hierarchical organisation, as seen in the graph layouts and the multi-level Sankey diagram.
It is important to remark that, although the Markov time acts as a natural resolution parameter from finer to coarser partitions, our process of optimisation does not impose any hierarchical structure \emph{a priori}. Hence the observed consistency of communities across level is intrinsic to the data and suggests the existence of content clusters that naturally integrate with each other as sub-themes of larger thematic categories. The detection of intrinsic scales within the graph provided by MS thus enables us to obtain clusters of records with high content similarity at different levels of granularity. This capability can be used by practitioners to tune the level of description to their specific needs.

\subsection*{Interpretation of MS communities: content and \textit{a posteriori} comparison with hand-coded categories}

\begin{figure}[htb]
\includegraphics[width=.97\linewidth]{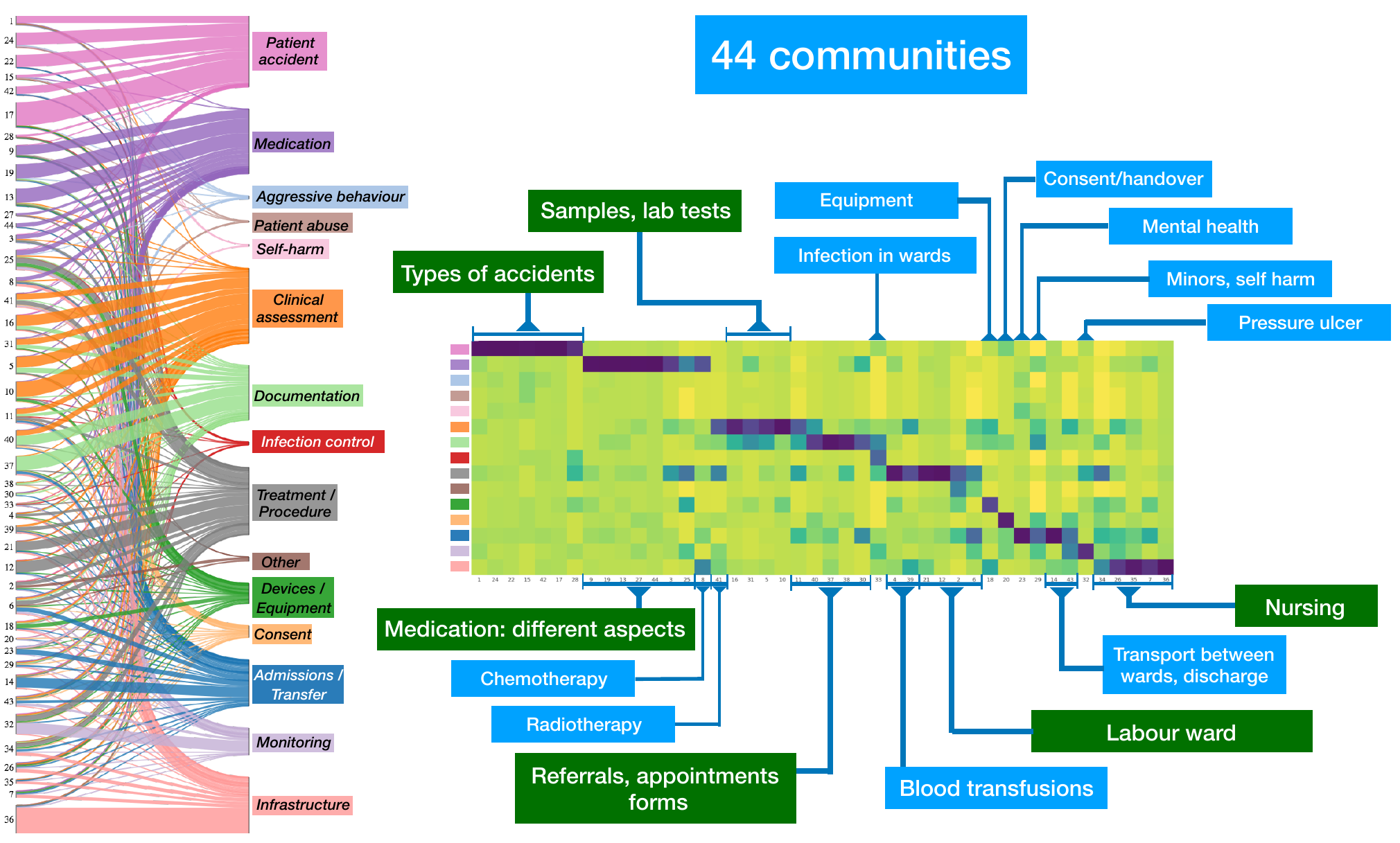}
\caption{Summary of the 44-community found with the MS algorithm in an unsupervised manner directly from the text of the incident reports, as seen in Figure~\ref{fig:MS}. To interpret the 44 content communities, we have compared them \textit{a posteriori} to the 15 external, hand-coded categories (indicated by names and colours). This comparison is presented in two equivalent ways: through a Sankey diagram showing the correspondence between categories and communities (left); and through a normalised contingency table based on z-scores (right). The communities have been assigned a content label based on their word clouds presented in Figure~\ref{fig:44comms_SI} in the SI.
}
\label{fig:44comms}
\end{figure}

\begin{figure}[htb]
\includegraphics[width=.97\linewidth]{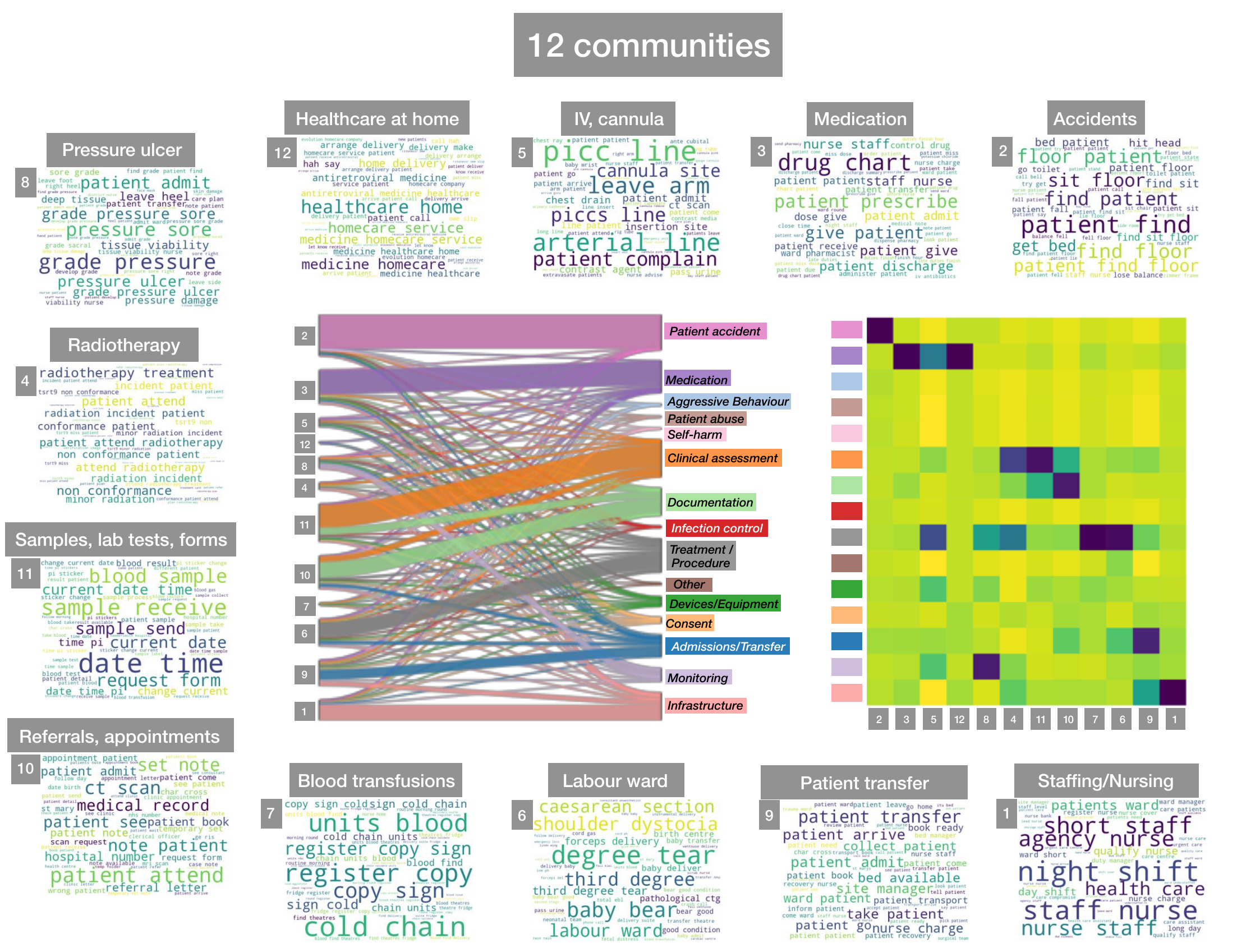}
\caption{Analysis of the results of the 12-community partition of documents obtained by MS based on their text content and their correspondence to the external categories. Some communities and categories are clearly matched while other communities reflect strong medical content.
}
\label{fig:12comms}
\end{figure}

\begin{figure}[p]
\includegraphics[width=.8\linewidth]{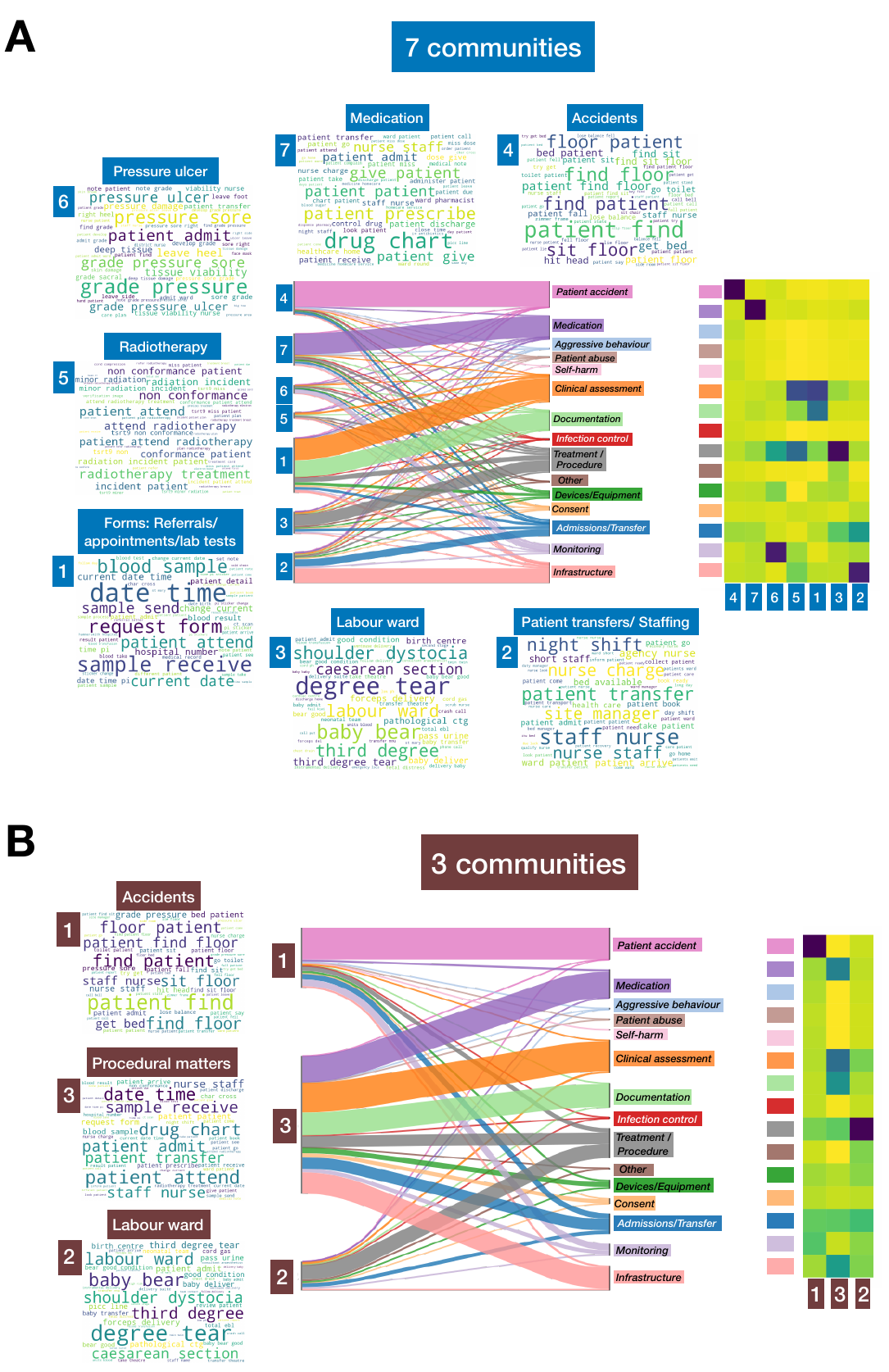} 
\centering
\caption{
Results for the coarser MS partitions of the document similarity graph into: (A) 7 communities and (B) 3 communities, showing in each case their correspondence to the external hand-coded categories. Some of the MS communities with strong medical content (e.g., labour ward, radiotherapy, pressure ulcer) remain separate in our content-driven, unsupervised clustering and are not integrated with other procedural records due to their semantic distinctiveness even to this coarsest level of clustering.   
}
\label{fig:7_3comms}
\end{figure}

To ascertain the relevance of the different layers of content clusters found in the MS analysis, we examined in detail the five levels of resolution presented in Figure~\ref{fig:MS}. For each level, we prepared word clouds (lemmatised for increased intelligibility), as well as a Sankey diagram and a contingency table linking content clusters (i.e., graph communities) with the hand-coded categories externally assigned by an operator. We note again that this comparison was only done \textit{a posteriori}, i.e., the external categories were not used in our text analysis. The results are shown in Figures~\ref{fig:44comms}--\ref{fig:7_3comms} (and Supplementary Figures~\ref{fig:44comms_SI}--\ref{fig:17comms_SI}) for all levels.

The partition into 44 communities presents content clusters with well-defined characterisations, as shown by the Sankey diagram and the highly clustered structure of the contingency table (Figure~\ref{fig:44comms}). The content labels for the communities were derived by us from the word clouds presented in detail in the Supplementary Information (Fig.~\ref{fig:44comms_SI} in the SI).
Compared to the 15 hand-coded categories, this 44-community partition provides finer groupings of records with several clusters corresponding to sub-themes or more specific sub-classes within large, generic hand-coded categories. 
This is apparent in the external classes `Accidents', `Medication', `Clinical assessment', `Documentation' and `Infrastructure', where a variety of subtopics are identified corresponding to meaningful subclasses (see Fig.~\ref{fig:44comms_SI} for details).
In other cases, however, the content clusters cut across the external categories, or correspond to highly specific content. Examples of the former are the content communities of records from \textit{labour ward, chemotherapy, radiotherapy and infection control}, whose reports are grouped coherently based on content by our algorithm, yet belong to highly diverse external classes. 
At this level of resolution, our algorithm also identified highly specific topics as separate content clusters. These include \textit{blood transfusions, pressure ulcer, consent, mental health}, and \textit{child protection}.

We have studied two levels of resolution where the number of communities (12 and 17) is close to that of hand-coded categories (15). The results of the 12-community partition are presented in Figure~\ref{fig:12comms} (see Figure~\ref{fig:17comms_SI} in the SI for the slightly finer 17-community partition).
As expected from the quasi-hierarchical nature of our multi-resolution analysis, we find that some of the communities in the 12-way partition emerge from consistent aggregation of smaller communities in the 44-way partition. In terms of topics, this means that some of the sub-themes observed in Figure~\ref{fig:44comms} are merged into a more general topic. This is apparent in the case of \textit{Accidents}: seven of the communities in the 44-way partition become one larger community (community 2 in Fig.~\ref{fig:12comms}), which has a specific and complete identification with the external category `Patient accidents'. A similar phenomenon is seen for the \textit{Nursing} community (community 1) which falls completely under the external category `Infrastructure'. The clusters related to `Medication' similarly aggregate into a larger community (community 3), yet there still remains a smaller, specific community related to \textit{Homecare medication} (community 12) with distinct content.

Other communities strand across a few external categories. This is clearly observable in communities 10 and 11 (\textit{Samples/ lab tests/forms} and \textit{Referrals/appointments}), which fall naturally across the external categories `Documentation' and `Clinical Assessment'. Similarly, community 9 (\textit{Patient transfers}) sits across the `Admission/Transfer' and `Infrastructure' external categories, due to its relation to nursing and other physical constraints. The rest of the communities contain a substantial proportion of records that have been hand-classified under the generic `Treatment/Procedure' class; yet here they are separated into groups that retain medical coherence, i.e., they refer to medical procedures or processes, such as \textit{Radiotherapy} (Comm.~4), \textit{Blood transfusions} (Comm.~7), \textit{IV/cannula} (Comm.~5), \textit{Pressure ulcer} (Comm.~8), and the large community \textit{Labour ward} (Comm.~6).

The high specificity of the \textit{Radiotherapy}, \textit{Pressure ulcer} and \textit{Labour ward} communities means that they are still preserved as separate groups on the next level of coarseness given by the 7-way partition (Figure~\ref{fig:7_3comms}A). The mergers in this case lead to a larger communities referring to \textit{Medication}, \textit{Referrals/Forms} and \textit{Staffing/Patient transfers}. Figure~\ref{fig:7_3comms}B shows the final level of agglomeration into 3 communities: a community of records referring to accidents; another community broadly referring to procedural matters (referrals, forms, staffing, medical procedures) cutting across many of the external categories; and the labour ward community still on its own as a subgroup of incidents with distinctive content. 

\begin{figure}[p]
\includegraphics[width=1.29\linewidth,angle=90]{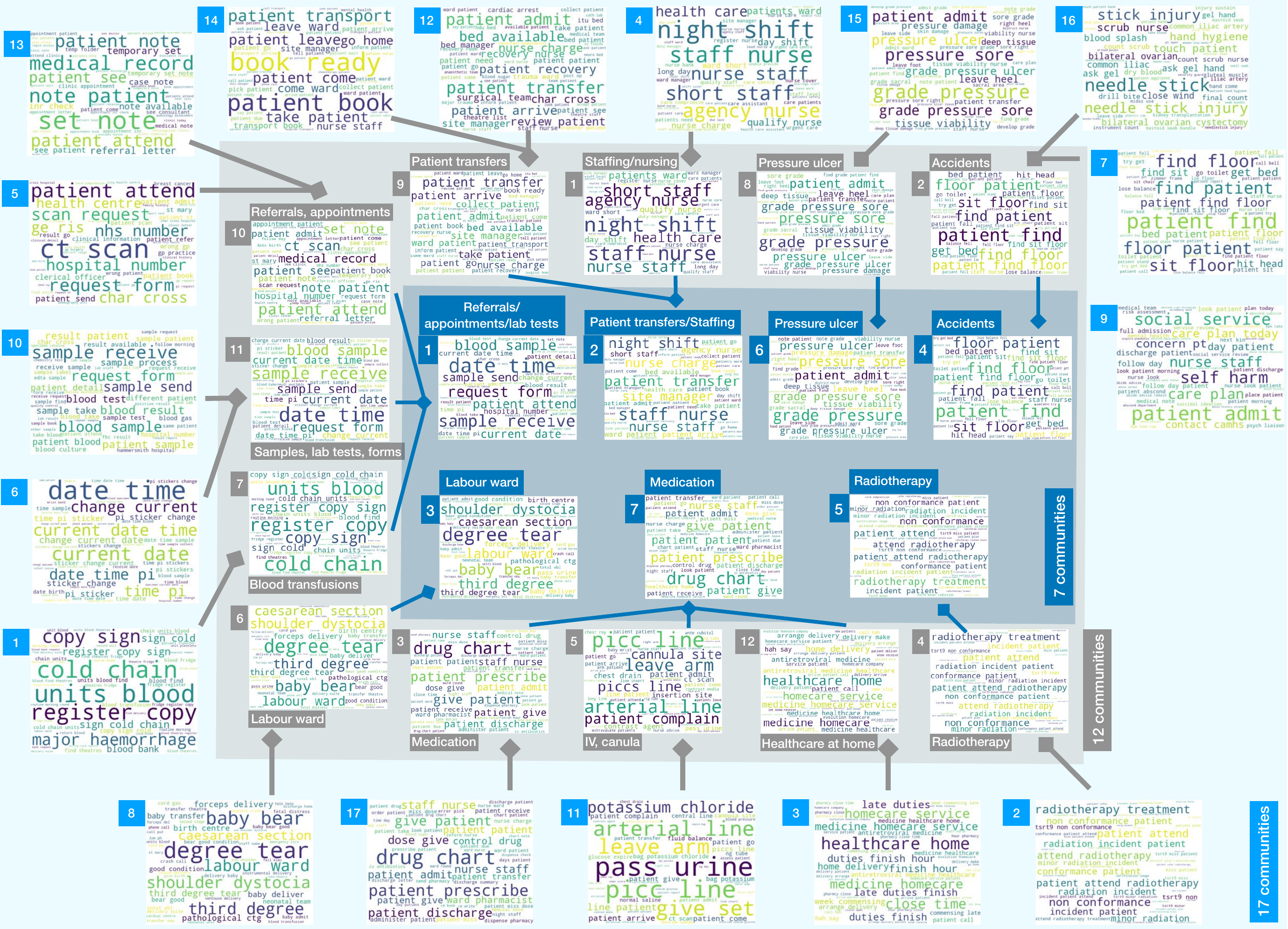}
\centering
\caption{
The word clouds of the partitions into 17, 12 and 7 communities show a multi-resolution coarsening in the content descriptive power mirroring the multi-level, quasi-hierarchical community structure found in the document similarity graph.
}
\label{fig:17_12_7_text}
\end{figure}

This process of agglomeration of content, from sub-themes into larger themes, as a result of the multi-scale hierarchy of graph partitions obtained with MS is shown explicitly with word clouds in Figure~\ref{fig:17_12_7_text} for the 17, 12 and 7-way partitions.

\subsection*{Robustness of the results and comparison with other methods}
\label{sec:comparisons}

Our framework consists of a series of steps for which there are choices and alternatives. Although it is not possible to provide comparisons to the myriad of methods and possibilities available, we have examined quantitatively the robustness of the results to parametric and methodological choices in different steps of the framework: (i) the importance of using Doc2Vec embeddings instead of BoW vectors, (ii) the size of training corpus for Doc2Vec; (iii) the sparsity of the MST-kNN similarity graph construction.  
We have also carried out quantitative comparisons to other methods, including: (i) LDA-BoW, and (ii) clustering with other community detection methods.
We provide a brief summary here and additional material in the SI.

\paragraph*{Quantifying the importance of Doc2Vec compared to BoW:}
The use of fixed-sized vector embeddings (Doc2Vec) instead of standard bag of words (BoW) is an integral part of our pipeline. Doc2Vec produces lower dimensional vector representations (as compared to BoW) with higher semantic and syntactic content. It has been reported that Doc2Vec outperforms BoW representations in practical benchmarks of semantic similarity, as well as being less sensitive to hyper-parameters~\citep{dai2015document}.

\begin{figure}[htb]
\centering
\includegraphics[width=.8\linewidth,angle=0]{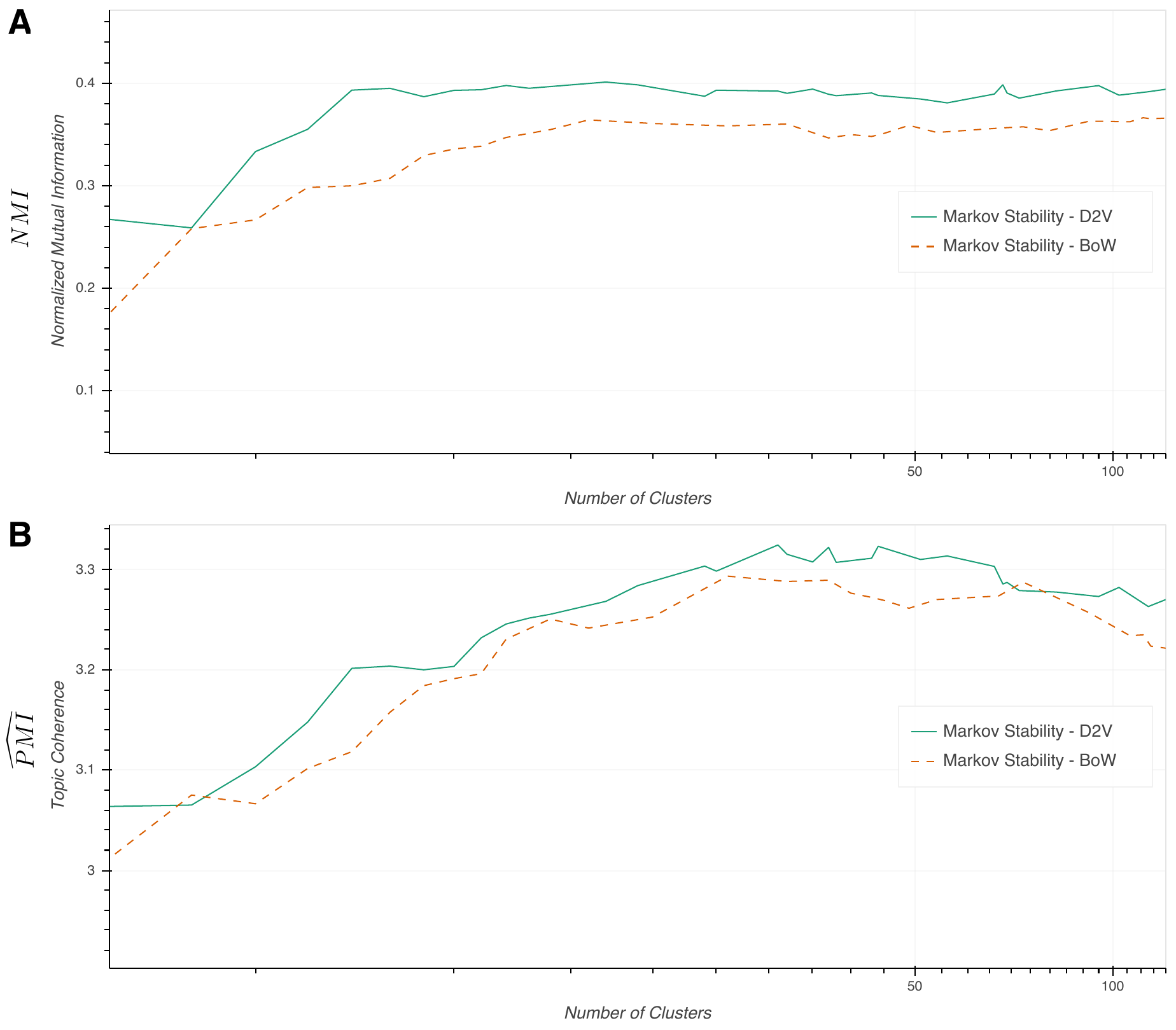}
\caption{
Comparison of MS applied to  Doc2Vec versus BoW (using TF-iDF) similarity graphs obtained after under the same graph constructions steps.
(A) Similarity against the externally hand-coded categories measured with $NMI$; (B) intrinsic topic coherence of the computed clusters measured with $\widehat{PMI}$.}
\label{fig:D2VvsBoW_MS}
\end{figure}

To quantify the improvement provided by Doc2Vec in our framework, we constructed a MST-kNN graph following the same steps but starting with TF-iDF vectors for each document. We then ran MS on this TF-iDF similarity graph, and compared the results to those obtained from the Doc2Vec similarity graph. Figure~\ref{fig:D2VvsBoW_MS} shows that the Doc2Vec version outperforms the BoW version across all resolutions in terms of both $NMI$ and $\widehat{PMI}$ scores.

\paragraph*{Robustness to the size of dataset to train Doc2Vec :}
As shown in Table~\ref{table:d2v}, we have tested the effect of the size of the training corpus on the Doc2Vec model. We trained Doc2Vec on two additional training sets of 1 million and 2 million records (randomly chosen from the full set of $\sim$13 million records). We then followed the same procedure to construct the MST-kNN similarity graph and carried out the MS analysis. 
The results, presented in Figure~\ref{fig:Corpus_size_SI} in the SI, show that the performance is affected only mildly by the size of the Doc2Vec training set.

\paragraph*{Robustness of the MS results to the level of sparsification:}
To examine the effect of sparsification in the graph construction, we have studied the dependence of quality of the partitions against the number of neighbours, $k$, in the MST-kNN graph. Our numerics, shown in Figure~\ref{fig:k_values_SI} in the SI, indicate that both the $NMI$ and $\widehat{PMI}$ scores of the MS clusterings reach a similar level of quality for values of $k$ above 13-16, with minor improvement after that. Hence our results are robust to the choice of $k$, provided it is not too small. Due to computational efficiency, we thus favour a relatively small $k$, but not too small.

\paragraph*{Comparison of MS clustering to Latent Dirichlet Allocation with Bag-of-Words (LDA-BoW):}
We carried out a comparison with LDA, a widely used methodology for text analysis. A key difference between standard LDA and our MS method is the fact that a different LDA model needs to be trained separately for each number of topics pre-determined by the user. To offer a comparison across the methods, We obtained five LDA models corresponding to the five MS levels we considered in detail. The results in Table~\ref{table:MSvsLDA_all} show that MS and LDA give partitions that are comparably similar to the hand-coded categories (as measured with $NMI$), with some differences depending on the scale, whereas the MS clusterings have higher topic coherence (as given by $\widehat{PMI}$) across all scales.

\begin{table}[htb]
\begin{center}
\begin{tabular}{c||c|c||c|c|}
 & \multicolumn{2}{c||}{\begin{tabular}[c]{@{}c@{}}Similarity to hand-coded \\ categories  ($NMI$)\end{tabular}}
 & \multicolumn{2}{c|}{\begin{tabular}[c]{@{}c@{}} Topic Coherence \\ ($\widehat{PMI}$)\end{tabular}} \\ \hline
 \multicolumn{1}{c||}{\textbf{\begin{tabular}[c]{@{}c@{}}No. of\\ topics/clusters\end{tabular}}} & 
\multicolumn{1}{c|}{LDA} & \multicolumn{1}{c||}{MS} & 
\multicolumn{1}{c|}{LDA} & \multicolumn{1}{c|}{MS} \\ 
\hline
3 	& \textbf{0.311} 	& 0.267 			& 2.991	 	& \textbf{3.033}		\\ \hline
7 	& \textbf{0.409} 	& 0.393 			& 3.218	 	& \textbf{3.303}		\\ \hline
12	& 0.361 			& \textbf{0.398} 	& 3.270		& \textbf{3.517}		\\ \hline
17	& 0.390 			& \textbf{0.401} 	& 3.419		& \textbf{3.457}		\\ \hline
44	& \textbf{0.395} 	& 0.388 			& 3.549		& \textbf{3.716}		\\ \hline
\end{tabular}
\caption{Scores for similarity to hand-coded categories ($NMI$) and topic coherence ($\widehat{PMI}$) for the five MS resolutions highlighted in the main text and their corresponding LDA models.}
\label{table:MSvsLDA_all}
\end{center}
\end{table}

To give an indication of the \emph{computational cost}, we ran both methods on the same servers. Our method takes approximately 13 hours in total to compute both the Doc2Vec model on 13 million records (11 hours) and the full MS scan with 400 partitions across all resolutions (2 hours). The time required to train just the 5 LDA models on the same corpus amounts to 30 hours (with timings ranging from $\sim$2 hours for the 3 topic LDA model to 12.5 hours for the 44 topic LDA model).

This comparison also highlights the conceptual difference between our multi-scale methodology and LDA topic modelling. While LDA computes topics at a pre-determined level of resolution, our method obtains partitions at all resolutions in one sweep of the Markov time, from which relevant partitions are chosen based on their robustness. However, the MS partitions at all resolutions are available for further investigation if so needed.

\paragraph*{Comparison of MS to other partitioning and community detection algorithms:}
We have used several algorithms readily available in code libraries (i.e., the iGraph module for Python) to cluster/partition the same kNN-MST graph. Figure~\ref{fig:commDetvsMS_SI} in the SI shows the comparison against several well-known partitioning methods (Modularity Optimisation~\citep{modularity_igraph}, InfoMap~\citep{infomap_EPJS},  Walktrap~\citep{walktrap}, Label Propagation~\citep{labelprop}, and Multi-resolution Louvain~\citep{louvain}) which give just one partition (or two in the case of the Louvain implementation in iGraph) into a particular number of clusters, in contrast with our multiscale MS analysis. 
Our results show that MS provides improved or equal results to other graph partitioning methods for both $NMI$ and $\widehat{PMI}$ across all scales. Only for very fine resolution with more than 50 clusters, Infomap, which partitions graphs into small clique-like 
subgraphs~\citep{Schaub2012ZoomingLens,schaub2012encoding}, provides a slightly improved $NMI$ for that particular scale. Therefore, MS allows us to find relevant, yet high quality clusterings across all scales by sweeping the Markov time parameter.

\section*{Discussion}

This work has applied a multiscale graph partitioning algorithm (Markov Stability) to extract content-based clusters of documents from a textual dataset of healthcare safety incident reports in an unsupervised manner at different levels of resolution. The method uses paragraph vectors to represent the records and obtains an ensuing similarity graph of documents constructed from their content. 
The framework brings the advantage of multi-resolution algorithms capable of capturing clusters without imposing \textit{a priori} their number or structure. Since different levels of resolution of the clustering can be found to be relevant, the practitioner can choose the level of description and detail to suit the requirements of a specific task.

Our \textit{a posteriori} analysis evaluating the similarity against the hand-coded categories and the intrinsic topic coherence of the clusters showed that the method performed well in recovering meaningful categories. The clusters of content capture topics of medical practice, thus providing complementary information to the externally imposed classification categories.
Our analysis shows that some of the most relevant and persistent communities emerge because of their highly homogeneous medical content, although they are not easily mapped to the standardised external categories. This is apparent in the medically-based content clusters associated with \textit{Labour ward}, \textit{Pressure ulcer}, \textit{Chemotherapy}, \textit{Radiotherapy}, among others, which exemplify the alternative groupings that emerge from free text content.

The categories in the top level (Level 1) of the pre-defined classification hierarchy are highly diverse in size (as shown by their number of assigned records), with large groups such as `Patient accident', `Medication', `Clinical assessment', `Documentation', `Admissions/Transfer' or `Infrastructure' alongside small, specific groups such as `Aggressive behaviour', `Patient abuse', `Self-harm' or `Infection control'. Our multi-scale partitioning finds corresponding groups in content across different levels of resolution, providing additional subcategories with medical detail within some of the large categories (as shown in Fig.~\ref{fig:44comms}~and~\ref{fig:44comms_SI}). 
An area of future research will be to confirm if the categories found by our analysis are consistent with a second level in the hierarchy of external categories (Level 2, around 100 categories) that is used less consistently in hospital settings. 
The use of content-driven classification of reports could also be important within current efforts by the World Health Organisation (WHO) under the framework for the International Classification for Patient Safety (ICPS)~\citep{who_ICPS}
to establish a set of conceptual categories to monitor, analyse and interpret information to improve patient care.

One of the advantages of a free text analytical approach is the provision, in a timely manner, of an intelligible description of incident report categories derived directly from the rich description in the 'words' of the reporter themselves. 
The insight from analysing the free text entry of the person reporting could play a valuable role and add rich information than would have otherwise been obtained from the existing approach of pre-defined classes. 
Not only could this improve the current state of play where much of the free text of these reports goes unused, but it avoids the fallacy of assigning incidents to a pre-defined category that, through a lack of granularity, can miss an important opportunity for feedback and learning. The nuanced information and classifications extracted from free text analysis thus suggest a complementary axis to existing approaches to characterise patient safety incident reports.

Currently, local incident reporting system are used by hospitals to submit reports to the NRLS and require risk managers to improve data quality of reports, due to errors or uncertainty in categorisation from reporters, before submission. The application of free text analytical approaches, like the one we have presented here, has the potential to free up risk managers time from labour-intensive tasks of classification and correction by human operators, instead for quality improvement activities derived from the intelligence of the data itself. Additionally, the method allows for the discovery of emerging topics or classes of incidents directly from the data when such events do not fit the pre-assigned categories by using projection techniques alongside methods for anomaly and innovation detection.

In ongoing work, we are currently examining the use of our characterisation of incident reports to enable comparisons across healthcare organisations and also to monitor their change over time. This part of ongoing research requires the quantification of in-class text similarities and to dynamically manage the embedding of the reports through updates and recalculation of the vector embedding. Improvements in the process of robust graph construction are also part of our future work. Detecting anomalies in the data to decide whether newer topic clusters should be created, or providing online classification suggestions to users based on the text they input are some of the improvements we aim to add in the future to aid with decision support and data collection, and to potentially help fine-tune some of the predefined categories of the external classification.


\theendnotes

\section*{Availability of Data and Materials}

The dataset in this work is managed by the Big Data and Analytics Unit (BDAU), Imperial College London, and consists of incident reports submitted to the NRLS. Analysis of the data was undertaken within the Secure Environment of the BDAU. Due to its nature, we cannot publicise any part of the dataset, beyond that already provided within this manuscript. No individual identifiable patient information is disclosed in this work. Only aggregated information is used to describe the clusters.

\section*{Competing interests}

The authors declare that they have no competing interests.

\section*{List of abbreviations}

\textbf{NHS:} National Health Service;
\textbf{NRLS:} National Reporting and Learning System;
\textbf{BoW:} Bag of Words;
\textbf{LDA:} Latent Dirichlet Allocation;
\textbf{Doc2Vec:} Document to Vector;
\textbf{MST:} Minimum Spanning Tree;
\textbf{kNN:} k-Nearest Neighbours;
\textbf{MS:} Markov Stability;
\textbf{NLTK:} Natural Language Toolkit;
\textbf{TF-iDF:} Term Frequency - inverse Document Frequency;
\textbf{PV:} Paragraph Vectors; 
\textbf{DBOW:} Distributed Bag of Words;
\textbf{VI:} Variation of Information;
\textbf{NMI:} Normalised Mutual Information;
\textbf{PMI:} Pairwise Mutual Information.

\section*{Authors' contributions}

MTA conducted the computational research. MTA and MB analysed the data and designed the computational framework. MB, EM and SNY conceived the study. All authors wrote the manuscript.

\section*{Acknowledgements}

We thank Joshua Symons for help with accessing the data. We also thank Elias Bamis, Zijing Liu and Michael Schaub for helpful discussions.
This research was supported by the National Institute for Health Research (NIHR) Imperial Patient Safety Translational Research Centre and NIHR Imperial Biomedical Research Centre. The views expressed are those of the authors and not necessarily those of the NHS, the NIHR, or the Department of Health. 
All authors acknowledge funding from the EPSRC through award EP/N014529/1 funding the EPSRC Centre for Mathematics of Precision Healthcare. 

\section*{Authors' Information}

MTA is a PhD student at Imperial College London, Department of Mathematics. He holds an MSc degree in finance from Sabanci University and a BSc in Electrical and Electronics  Engineering from Bogazici University. EM is a Clinical Senior Lecturer in the Department of Surgery and Cancer  and Centre for Health Policy at Imperial College London and Transformation Chief Clinical Information Officer (Clinical Analytics and Informatics), ICHNT. SNY is a Professor of Theoretical Chemistry in the Department of Chemistry at Imperial College London and also with the EPSRC Centre for Mathematics of Precision Healthcare. MB is Professor of Mathematics and Chair in Biomathematics in the Department of Mathematics at Imperial College London, and Director of the EPSRC Centre for Mathematics of Precision Healthcare at Imperial.


\bibliographystyle{naturemag}

\newpage
\renewcommand{\figurename}{Figure}
\renewcommand{\thefigure}{S\arabic{figure}}
\renewcommand{\tablename}{Table}
\renewcommand{\thetable}{S\arabic{table}}
\renewcommand{\theequation}{S\arabic{equation}}

\renewcommand{\thesection}{S\arabic{section}}

\setcounter{figure}{0}
\setcounter{table}{0}
\setcounter{section}{0}

\section*{Additional Files}

\begin{figure}[h!]
\includegraphics[width=.97\linewidth]{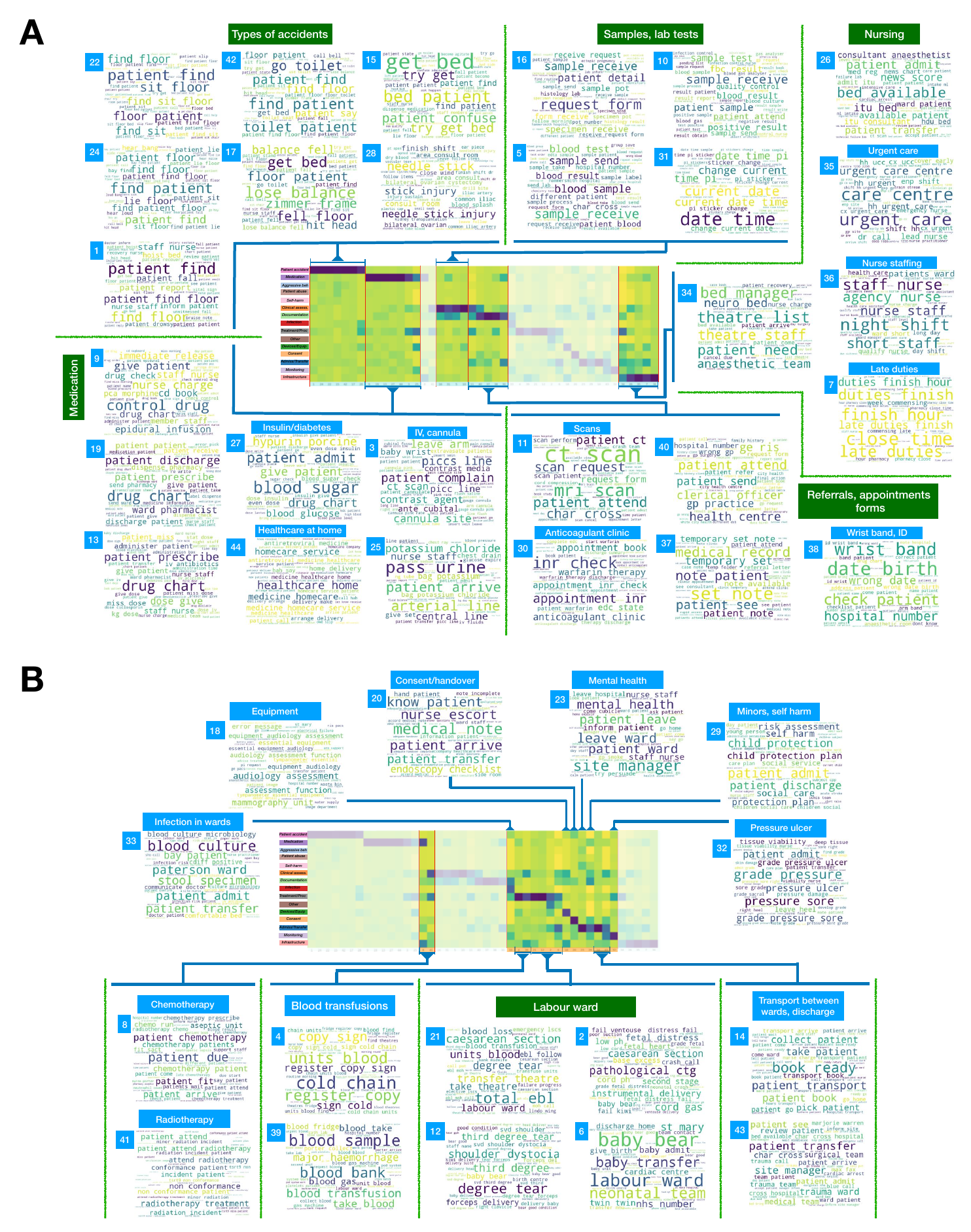}
\caption[Additional file 1 --- Word clouds for the 44 community partition]{Additional file 1 --- Word clouds for the 44 community partition \par
Word clouds of the 44-community partition showing the detailed content of the communities found. The word clouds are split into two subfigures (A) and (B) for ease of visualisation.
}
\label{fig:44comms_SI}
\end{figure}

\begin{figure}[h!]
\includegraphics[width=\linewidth]{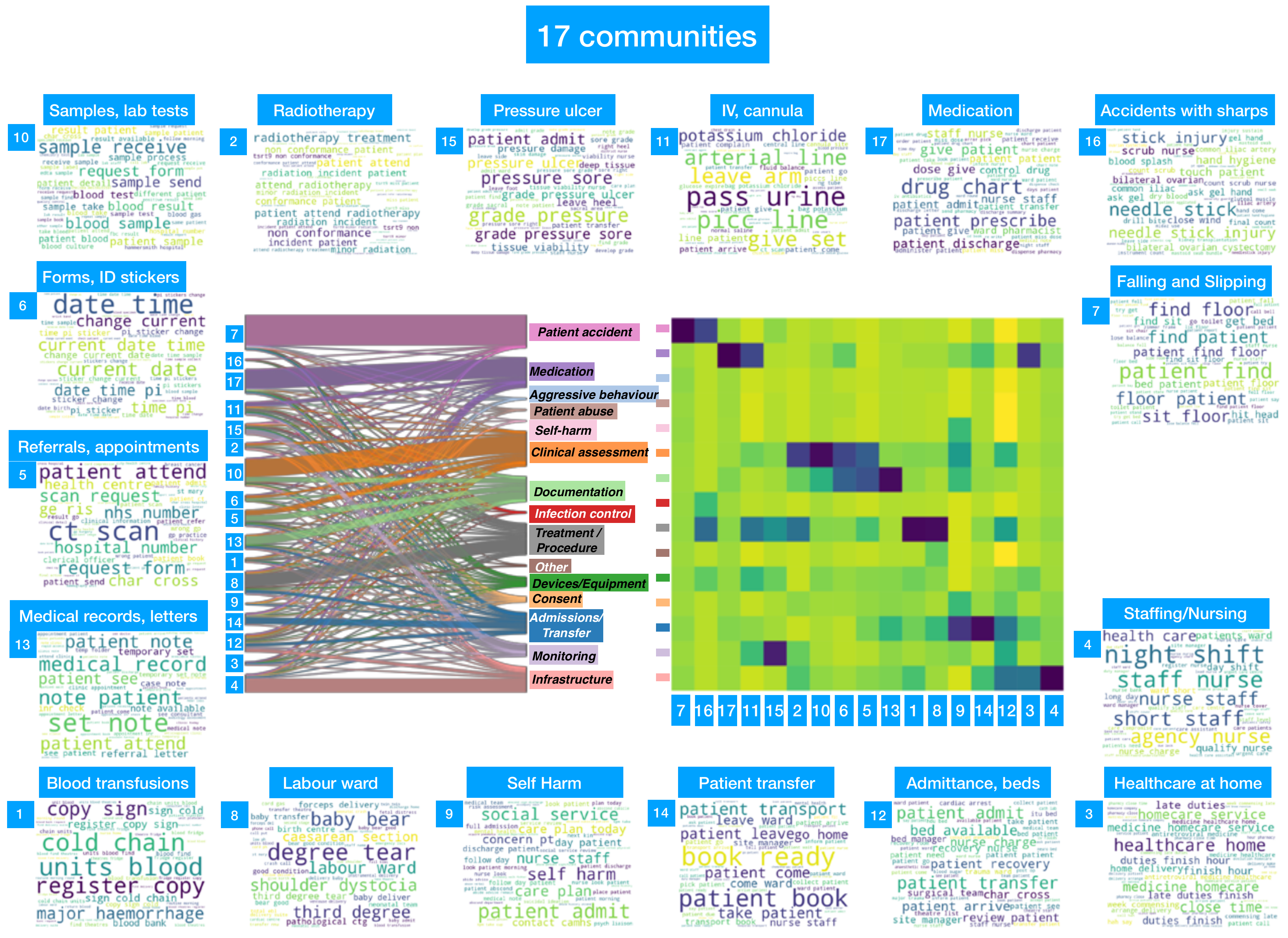}
\caption[Additional file 2 --- Word cloud and Sankey diagram for the 17 community partition]{Additional file 2 --- Word cloud and Sankey diagram for the 17 community partition \par 
Analysis of the results of the 17-community MS partition and their correspondence to the external categories. Compared to the 12-way partition in the main text, this slightly finer partition shows some communities with more detailed medical content, as shown in Figure~\ref{fig:17_12_7_text}.
}
\label{fig:17comms_SI}
\end{figure}

\begin{figure}[p]
\includegraphics[width=\linewidth]{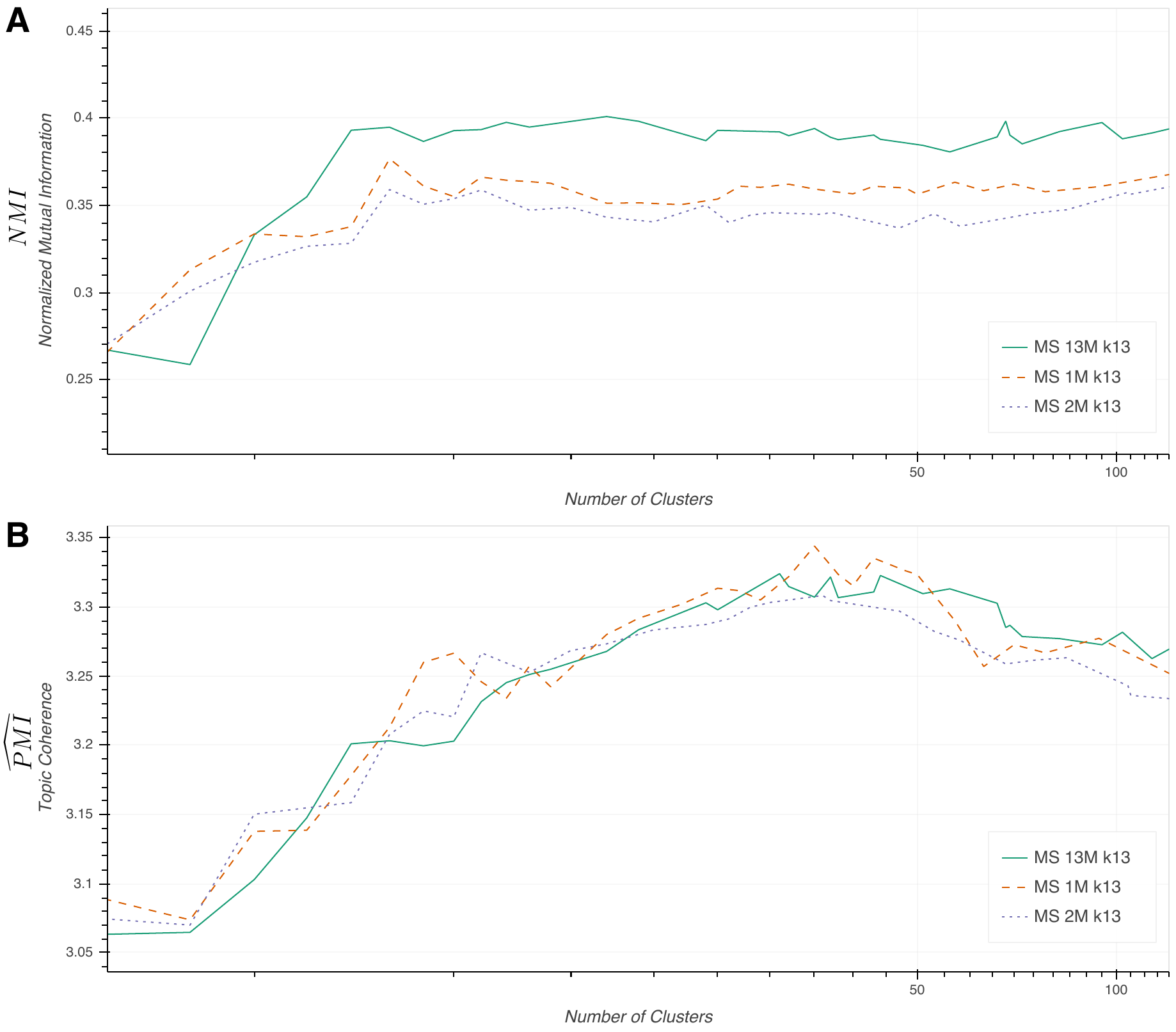}
 \caption[Additional file 3 --- Effect of the corpus size]{Additional file 3 --- Effect of the corpus size \par
 Evaluating the effect of the size of the training corpus
 (A) Similarity to hand-coded categories (measured with $NMI$) and (B) Topic Coherence score (measured with $\widehat{PMI}$) of the MS clusterings obtained across all Markov times when applied to the similarity graph of documents obtained from three different Doc2Vec embeddings trained on: 1 million records, 2 million records, and the full set of 13 million records. The corpus size does not affect the results.
 }
\label{fig:Corpus_size_SI}
\end{figure}

\begin{figure}[p]
\includegraphics[width=\linewidth]{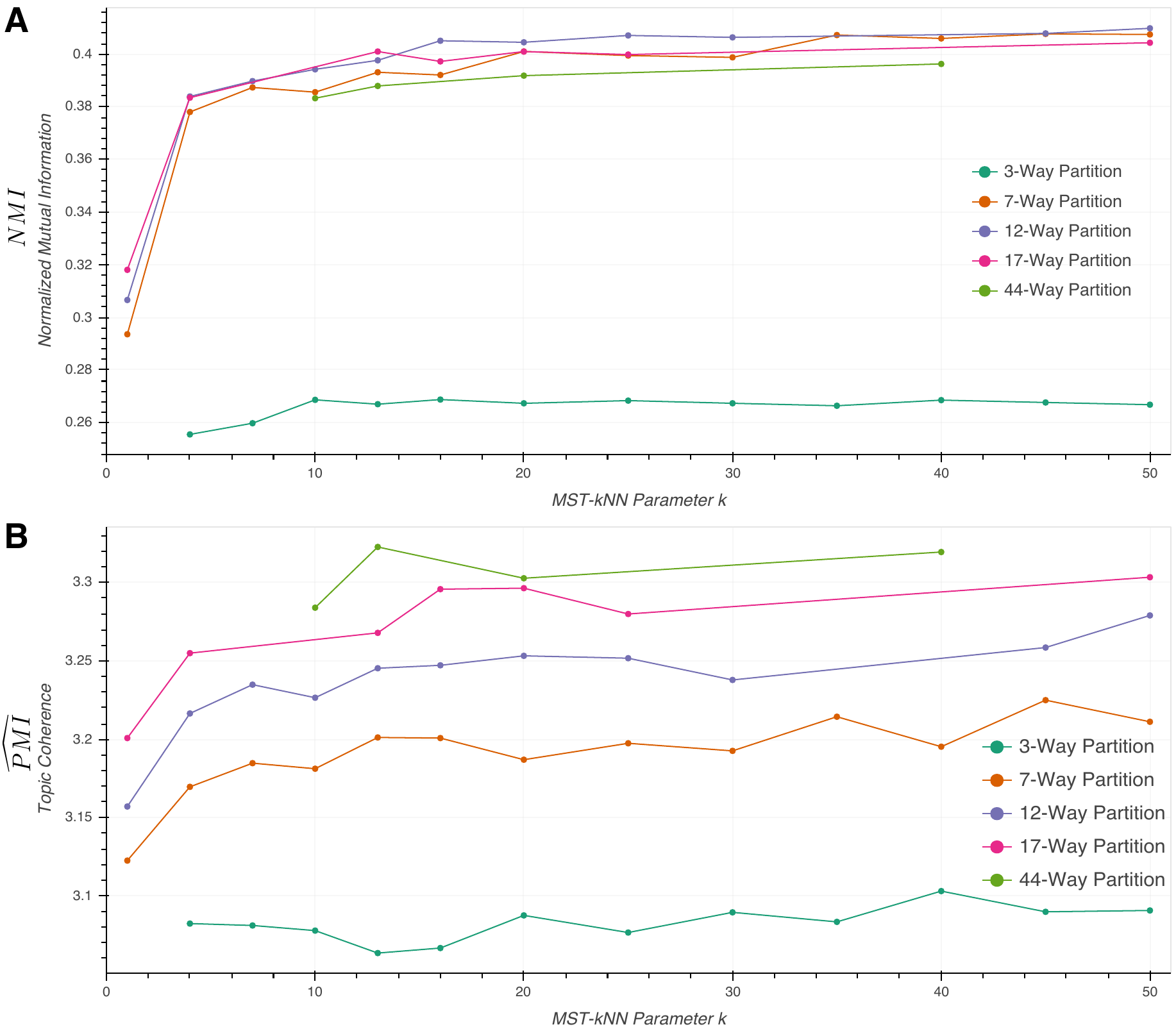}
\caption[Additional file 4 --- Effect of the sparsification]{Additional file 4 --- Effect of the sparsification \par
Comparison of MS applied to MST-kNN similarity graphs with increasing $k$. 
(A) Similarity against the externally hand-coded categories measured with $NMI$; (B) Intrinsic topic coherence of the computed clusters measured with $\widehat{PMI}$.
}
\label{fig:k_values_SI}
\end{figure}

\begin{figure}[p]
\includegraphics[width=\linewidth]{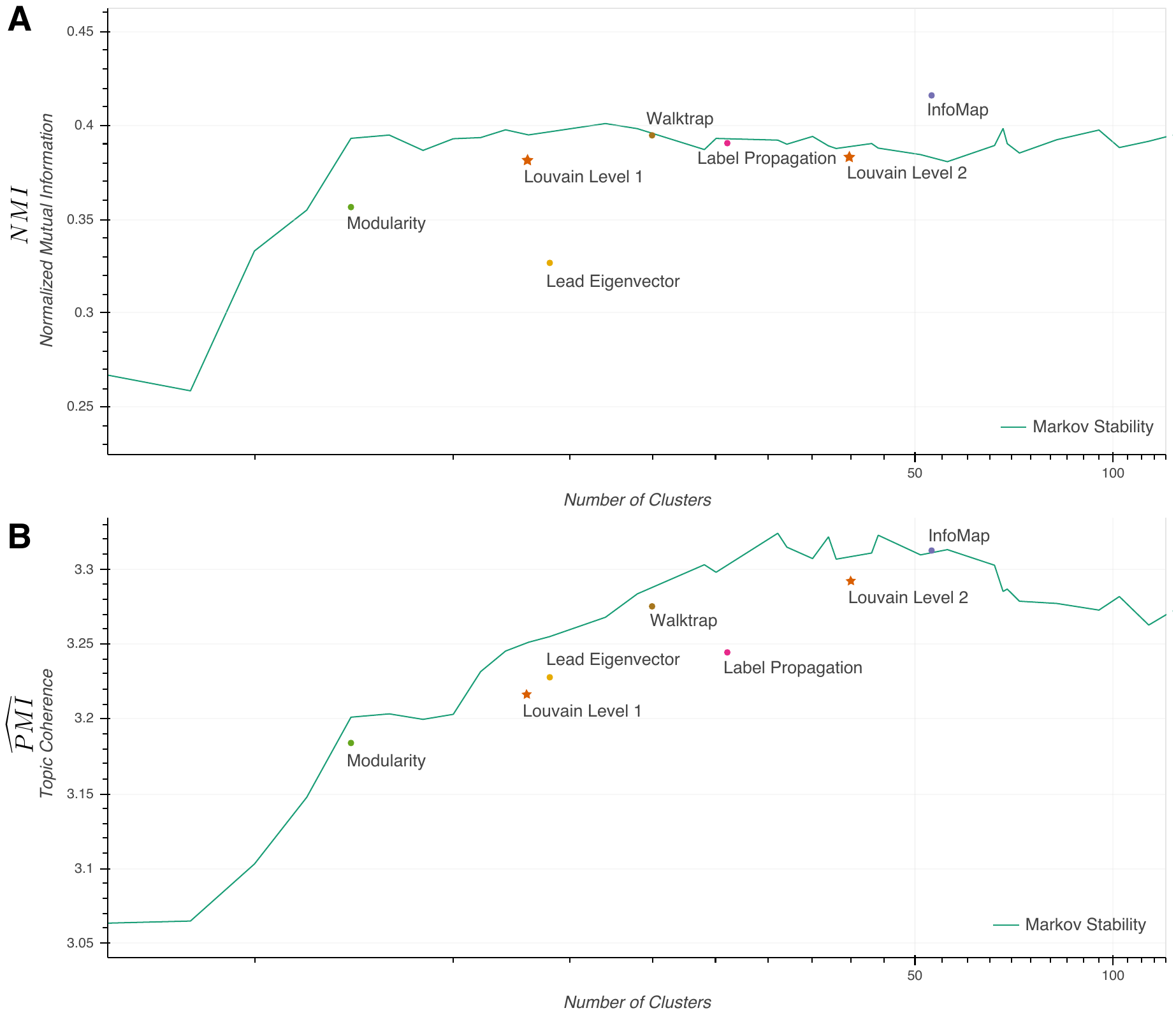}
\caption[Additional file 5 --- Comparison with other clustering methods]{Additional file 5 --- Comparison with other clustering methods \par
Comparison of MS results versus other common graph-based community detection or partitioning methods across all resolutions: (A) Similarity against the externally hand-coded categories measured with $NMI$; (B) intrinsic topic coherence of the computed clusters measured with $\widehat{PMI}$.  
}
\label{fig:commDetvsMS_SI}
\end{figure}

\end{document}